\documentclass[10pt,twocolumn,letterpaper]{article}

\usepackage[pagenumbers]{cvpr}

\usepackage{times}
\usepackage{epsfig}
\usepackage{url}
\usepackage{booktabs}
\usepackage{amsfonts}
\usepackage{nicefrac}
\usepackage{microtype}
\usepackage{xcolor}
\usepackage{colortbl}
\usepackage[export]{adjustbox}

\usepackage{graphicx}
\usepackage{tikz}
\usepackage{amsmath}
\usepackage{amssymb}
\usepackage{bbm}
\usepackage{bm}
\usepackage{booktabs}
\usepackage{alias}
\usepackage{lib}
\usepackage{cuted}
\usepackage{xcolor}
\usepackage{array}
\usepackage{multirow}
\newcolumntype{C}[1]{>{\centering\arraybackslash}p{#1}}
\usepackage{subcaption}
\usepackage{pifont}%

\definecolor{cvprblue}{rgb}{0.21,0.49,0.74}
\usepackage[pagebackref,breaklinks,colorlinks,allcolors=cvprblue]{hyperref}

\title{How Do I Do That? \\ Synthesizing 3D Hand Motion and Contacts for Everyday Interactions}

\author{Aditya Prakash\thanks{Part of the work was done during an internship at Microsoft}  $^{1}$
\quad \quad
Benjamin Lundell$^{2}$
\quad \quad
Dmitry Andreychuk$^{2}$
\quad \quad
David Forsyth$^{1}$ \\
Saurabh Gupta\thanks{Joint last authors, indicates equal contribution} $^{1}$
\quad \quad
Harpreet Sawhney\footnotemark[2] $^{2}$ \\
$^{1}$University of Illinois Urbana-Champaign \quad \quad $^{2}$Microsoft\\
{\tt \url{https://bit.ly/LatentAct}}
}

\begin{document}
\maketitle

\begin{strip}
    \vspace{-2cm}
    \centering
    \includegraphics[width=\linewidth]{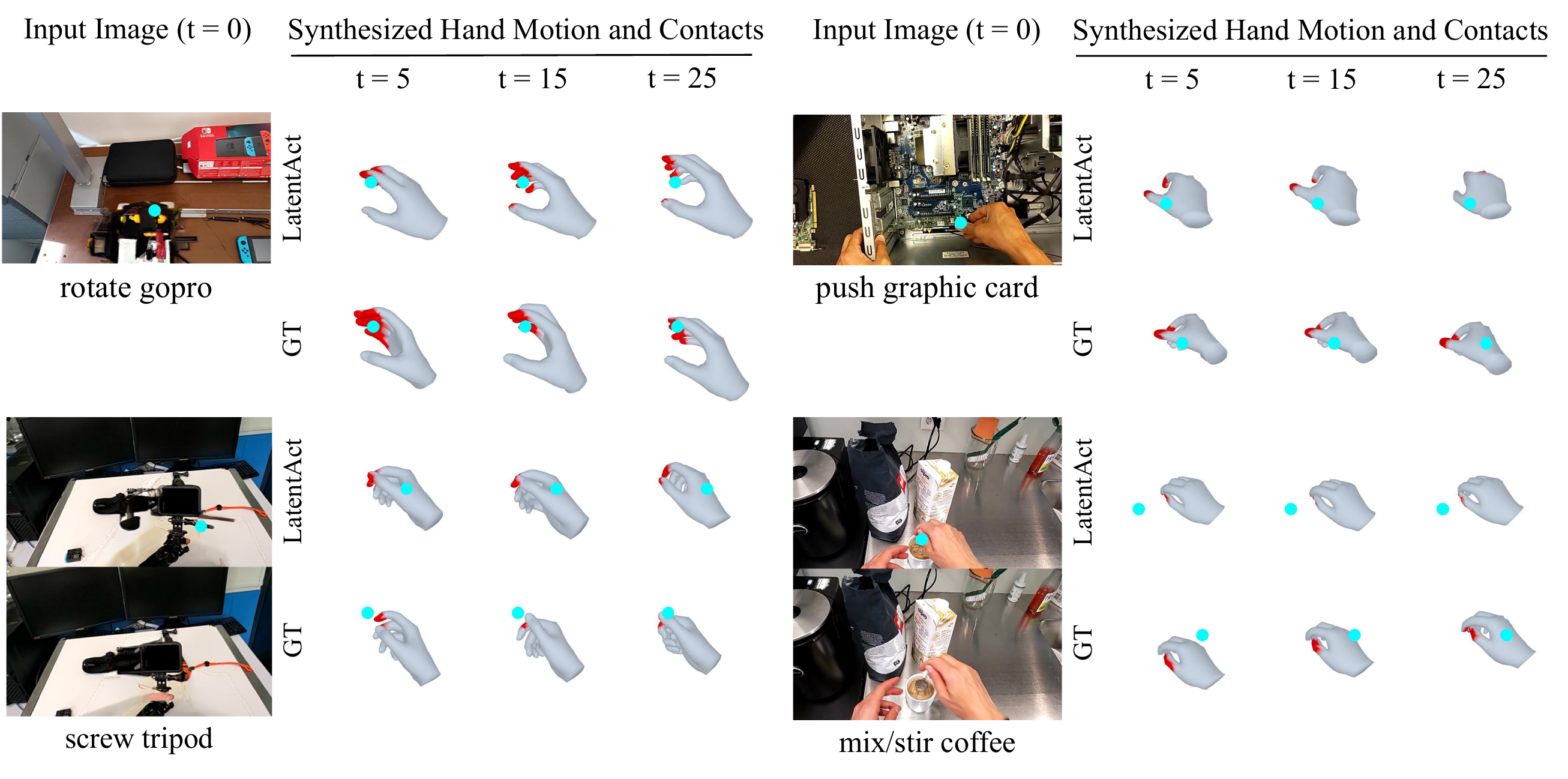}
    \vspace{-0.75cm}
    \captionof{figure}{\textbf{Interaction Trajectories}. We tackle the novel task of predicting future 3D hand poses \& {\color{red}{contact maps}}, i.e., interaction trajectories, from a single image showing the object, text describing the action and a 3D contact point on the object, in everyday activites. We show the trajectory predicted by our method (\ltf) \& ground truth (GT) for 3 future timesteps along with the {\color{cyan}{contact point}}. We consider 2 settings (top) Forecasting: single RGB view, action text \& 3D contact points as input, (bottom) Interpolation: goal image is also provided.}
    \figlabel{fig:teaser}
\end{strip}

\begin{abstract}
    We tackle the novel problem of predicting 3D hand motion and contact maps (or Interaction Trajectories) given a single RGB view, action text, and a 3D contact point on the object as input.
    Our approach consists of (1) Interaction Codebook: a VQVAE model to learn a latent codebook of hand poses and contact points, effectively tokenizing interaction trajectories, (2) Interaction Predictor: a transformer-decoder module to predict the interaction trajectory from test time inputs by using an indexer module to retrieve a latent affordance from the learned codebook. 
    To train our model, we develop a data engine that extracts 3D hand poses and contact trajectories from the diverse \holo dataset.
    We evaluate our model on a benchmark that is 2.5-10$\times$ larger than existing works, in terms of diversity of objects and interactions observed, and test for generalization of the model across object categories, action categories, tasks, and scenes.
    Experimental results show the effectiveness of our approach over transformer \& diffusion baselines across all settings.
\end{abstract}

\section{Introduction}
\label{sec:intro}

Hands interact with objects in diverse ways, often requiring varying skill levels to complete different tasks. Consider teaching someone to repair a bike. It is challenging to specify what part to manipulate~\cite{sodhi2013bethere}, what axis to flip a part over such that it inserts well, \etc. Instead, it is easier to directly show how the hand should move and what parts of it should make contact with the object over time to complete the task. In this work, we study this problem.  Given a single image of an object, a 3D contact point, and a text indicating the high-level action, we predict \textit{Interaction Trajectories}, the sequence of 3D hand poses and contact maps that specify the initial and evolving hand interactions (\Figref{fig:teaser}), interpretable by a human. We target interaction with everyday objects that are often small, thin, transparent, occluded, and deformable (\Figref{fig:teaser}) for which 3D object models may not be available. Our work is a first step towards learning a motion prior for diverse interactions in everyday activities, that can be combined with 3D object models in the future.

In the absence of a 3D model, explicitly specifying contact points in the scene grounds the hand-object interaction, especially when there are multiple objects \& 3D models are unavailable. This information is provided in the form of a 3D contact point. Moreover, we observe that while objects are diverse, hands interact with them in stylized ways~\cite{rosales2006combining}. There are only a few prototypical ways in which hands act in everyday interactions. To exploit this property, we learn a latent codebook of trajectories of 3D hand motions \& contact points (Interaction Codebook, \Secref{sec:latent}), effectively tokenizing \textit{interaction trajectories} using a VQVAE~\cite{Oord2017NEURIPS}. Intuitively, the codebook learns the prototypical motion involved in (say) screwing in a screw that can be repurposed into motion involved in (say) closing a bottle cap. The codebook is indexed using visual inputs at test time to retrieve the closest codebook entry (Indexer module, \Secref{sec:indexer}), and aligned with the scene using the 3D contact point \& visual context from the image input (transformer-decoder based Interaction Predictor, \Secref{sec:predictor}). We refer to our approach as {\em \ltf}.

Since existing HOI datasets with 3D annotations have constrained setups, we use egocentric videos from the \holo dataset, showing everyday interactions. We leverage recent advances in 3D hand pose estimation~\cite{Pavlakos2024CVPR} and segmentation in videos~\cite{Ravi2024ARXIV} to build a semi-automatic data engine (\Secref{sec:data}) to extract trajectories of 3D hand poses and contact maps from \holo for 800 tasks performed by ordinary users across 120 object categories and 24 action categories. Our setup is 2.5X-10X larger than existing works, \eg HOT3D~\cite{Banerjee2024ARXIV} (33 rigid objects),
ARCTIC~\cite{Fan2023CVPR} (10 articulated objects),
GRAB~\cite{taheri2020grab} (51 objects), HOI4D~\cite{Liu2022CVPR}
(16 object categories), in terms of diversity of objects and interactions observed in everyday activities. 

For evaluation, we focus on generalization to 4 aspects: novel object categories, action categories, tasks and scenes. We show results on 2 settings: (1) Forecasting: the model generated predictions conditioned on the textual description of the action, current image showing the object of interaction and the 3D contact point at the current timestep. (2) Interpolation: Here, we also provide the goal image, along with the previous inputs, showing the final state of interaction. For each of these settings, we consider 2 variants: (a) hand visible in the image, (b) hand is absent in the image (often the case at start of interactions). Across this large-scale experiment setting, we find our approach to generalize better than transformer \& diffusion baselines that directly learn to predict interaction trajectories from images (\Secref{sec:results}).

\section{Related Work}
\label{sec:related}

\boldparagraph{Generating HOI sequences}
Recent works in HOI~\cite{Christen2024ARXIV,Taheri2024THREEDV,Li2024WACV,Ghosh2023EUROGRAPHICS,Braun2023ARXIV,Zhang2023ARXIV,Ye2024CVPR} have studied generating hand-object motions in different settings, e.g. conditioned on textual descriptions~\cite{Christen2024ARXIV,Ghosh2023EUROGRAPHICS}, body poses~\cite{Braun2023ARXIV,Ghosh2023EUROGRAPHICS,Li2024WACV,Taheri2024THREEDV} for both rigid~\cite{Braun2023ARXIV,Christen2024ARXIV,Ghosh2023EUROGRAPHICS,Li2024WACV} and articulated objects~\cite{Zhang2023ARXIV,Cha2024CVPR}. While they focus on pick \& place tasks involving simple \& constrained interactions, we consider everyday activities, often involving small, thin, textureless, occluded, articulated, deformable and dynamic objects in contact for which obtaining 3D models is non-trivial. Existing works in this domain use either object-centric~\cite{Christen2024ARXIV,Cha2024CVPR} or hand-centric~\cite{Taheri2024THREEDV,Ye2024CVPR} frame to represent trajectories, this becomes a limitation in our setting since 3D objects are unknown and hands may not be present in the image (often the case at the start of the interaction). Instead, we focus on predicting future 3D hand poses \& contact maps from a single image showing the object, text describing the action \& a 3D contact point to localize the interaction, for diverse interactions in everyday activities.

\boldparagraph{HOI datasets} Existing works use datasets~\cite{Fan2023CVPR,hampali2020honnotate,kwon2021h2o,Hampali2022CVPR,Liu2022CVPR,huang2022reconstructing} collected in controlled settings which provide 3D ground truth using MoCap~\cite{Fan2023CVPR, taheri2020grab} or multi-camera setups~\cite{ohkawa2023assemblyhands, kwon2021h2o,hampali2020honnotate,Hampali2022CVPR,Liu2022CVPR,huang2022reconstructing}. However, the diversity of objects and interactions in these datasets is limited due to constrained nature of the capture setup. While recent datasets~\cite{darkhalil2022visor, shan2020understanding, cheng2023towards} have explored natural interactions in egocentric videos~\cite{damen2018scaling,grauman2022ego4d}, they only provide 2D annotations, \eg segmentation masks~\cite{darkhalil2022visor,cheng2023towards}, 2D bounding boxes~\cite{shan2020understanding} and grasp labels~\cite{cheng2023towards}. To mitigate this issue, we design a data engine to extract 3D hand poses and contact maps (\Secref{sec:data}) from egocentric videos showing everyday interactions. Specifically, we extract 3D trajectories from \holo for 800 tasks performed by ordinary users across 120 object categories and 24 action categories, which is 2.5X-10X larger than existing datasets~\cite{Banerjee2024ARXIV,taheri2020grab,Fan2023CVPR,Liu2022CVPR}.

\boldparagraph{HOI representation}
Hands are often represented using a parametric mesh model, e.g., MANO~\cite{Romero2017TOG}, along with the global translation and rotation of the wrist~\cite{Christen2024ARXIV,Ghosh2023EUROGRAPHICS, ye2022ihoi,Ye2023CVPR,Ye2024CVPR,Fan2023CVPR,prakash2023hoi,Prakash2024Handsb,Pavlakos2024CVPR}. Given an object mesh, its 6DoF pose can be denoted using translation \& rotation~\cite{Ghosh2023EUROGRAPHICS,Zhang2023ARXIV}, signed distance field~\cite{Ye2024CVPR} or using Basis Point Set (BPS~\cite{Prokudin2019ICCV}) distances~\cite{Christen2024ARXIV,Li2024WACV}. For hand-object contact, existing approaches typically consider a canonicalized representation with the hand~\cite{Ye2024CVPR,Cha2024CVPR} or object~\cite{Christen2024ARXIV} at the origin \& use relative distances~\cite{Christen2024ARXIV,Taheri2024THREEDV,Fan2023CVPR,Prakash2023Ambiguity,Fan2024ECCV} or a distance field~\cite{Ye2024CVPR} from hand joints to represent contact. Since 3D hand-object motion annotations are unavailable for diverse objects, e.g., thin, occluded, deformable, we instead focus on sequences of 3D hand poses and contact maps, \ie, interaction trajectories.

\boldparagraph{Affordances}
Gibson~\cite{Gibson1979} developed the concept of affordances as a set of functionalities that the environment furnishes to an agent, which can be a human, a robot, an animal, or hands. Affordances are typically represented as the contact regions on the objects~\cite{Bahl2023CVPR,Liu2022CVPR,Jiang2021ICCV,Li2023CVPR,Yang2023ICCV,taheri2020grab,chang2023neurips,Liu2023ICCV} \& the agent~\cite{Ye2024CVPR,Ye2023CVPR,Ye2023CVPR,Bahl2023CVPR,Jiang2021ICCV,Hassan2021CVPR,Tripathi2023ICCV,taheri2020grab,Srirama2024arxiv}. They have been explored in domains like robotics~\cite{Bahl2023CVPR,Srirama2024arxiv}, human interactions~\cite{Hassan2021CVPR,Tripathi2023ICCV,taheri2020grab,Xu2023ICCV} \& hand grasp synthesis~\cite{Ye2024CVPR,Ye2023CVPR,Jiang2021ICCV,taheri2020grab,Srirama2024arxiv}. Recent works have also proposed generative models for affordances synthesis~\cite{Ye2024CVPR,Ye2023CVPR,Xu2023ICCV}.

\section{Method}
\seclabel{sec:method}

Given a single RGB view, action text, and a 3D contact point as input, we focus on predicting 3D hand poses \& contact points, \ie, interaction trajectories (\Figref{fig:teaser}), for future timesteps.
Our approach consists of: (1) Interaction Codebook: learns a latent codebook of affordances using a VQVAE (\Figref{fig:LatentAct}), (2) Learned Indexer: maps the action text, image (showing an object) \& contact point inputs to the codebook indices to obtain the corresponding embeddings (\Figref{fig:predictor}), (3) Interaction Predictor: takes the queried embeddings along with text, image and contact point inputs to output the 3D interaction trajectory (\Figref{fig:predictor}).

\begin{figure}
    \centering
    \includegraphics[width=\columnwidth]{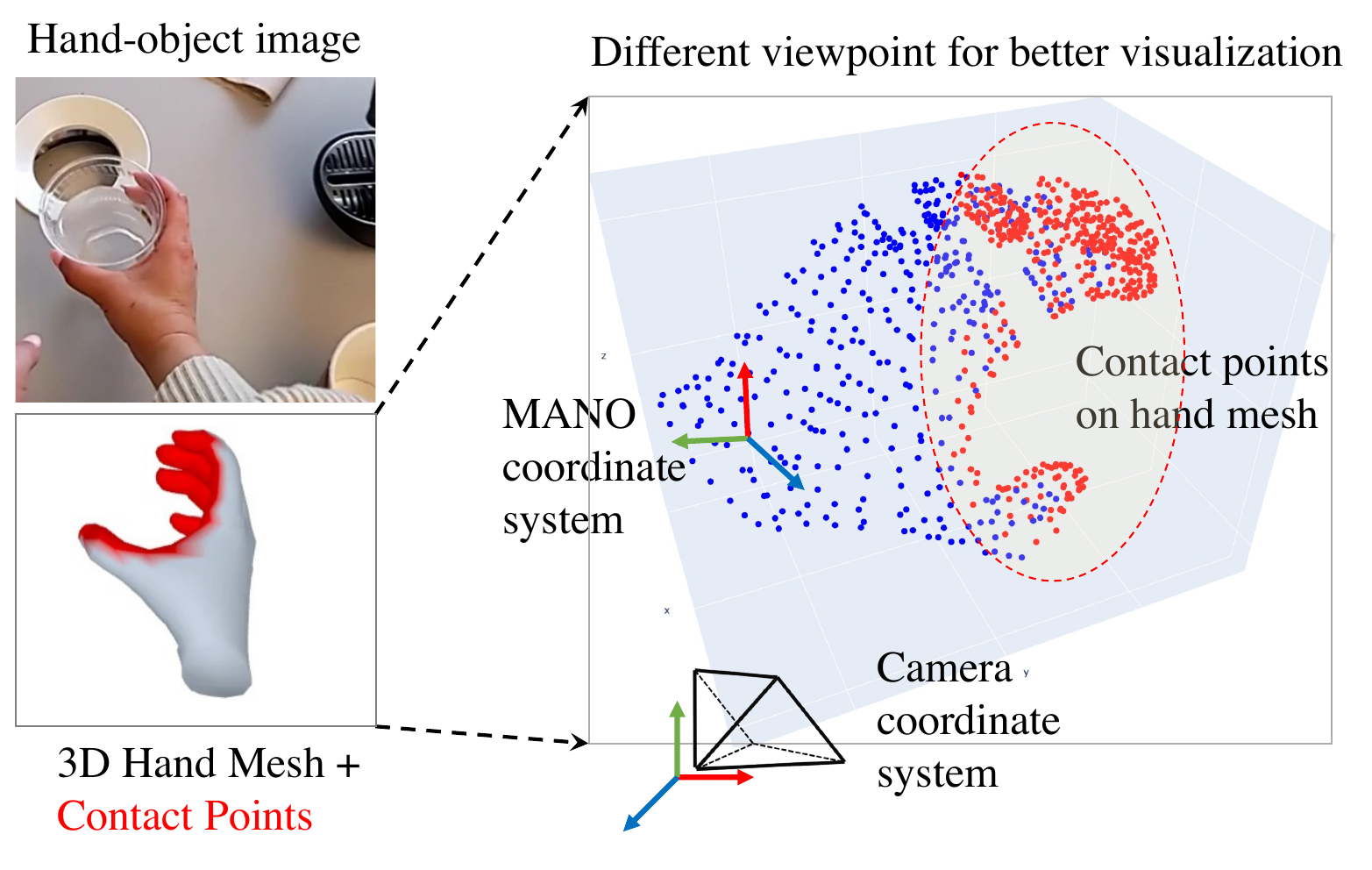}
    \caption{We represent {\color{red}{contact points}} as binary masks on the hand mesh vertices. The hand mesh is represented in the camera coordinate system, consisting of the local hand pose in the MANO~\cite{romero2017embodied} coordinate frame and a global transformation from the MANO frame to the camera frame. These hand poses and contact points over several timesteps form interaction trajectories.}
    \figlabel{fig:coordinate_frame}
    \vspace{-0.3cm}
\end{figure}

\begin{figure*}
    \begin{subfigure}{0.48\linewidth}
        \centering
        \includegraphics[width=\linewidth]{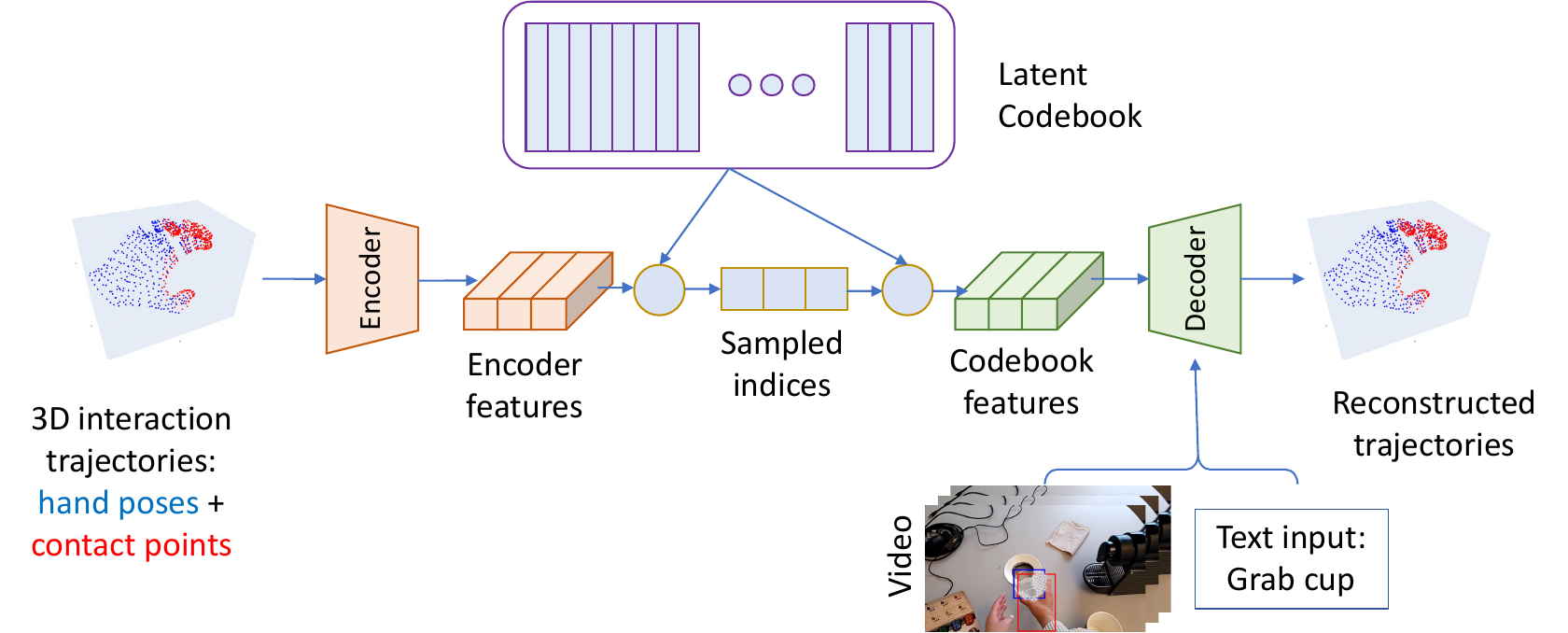}
        \caption{Interaction Codebook is a VQVAE model that learns a latent codebook of 3D interaction trajectories, consisting of a transformer encoder-decoder architecture. The encoder features are used to sample codebook indices and corresponding embeddings which are passed to a decoder that also takes in the video input and text describing the action and reconstructs the 3D interaction trajectories.}
        \figlabel{fig:LatentAct}
    \end{subfigure}
    \hfill
    \begin{subfigure}{0.48\linewidth}
        \centering
        \includegraphics[width=\linewidth]{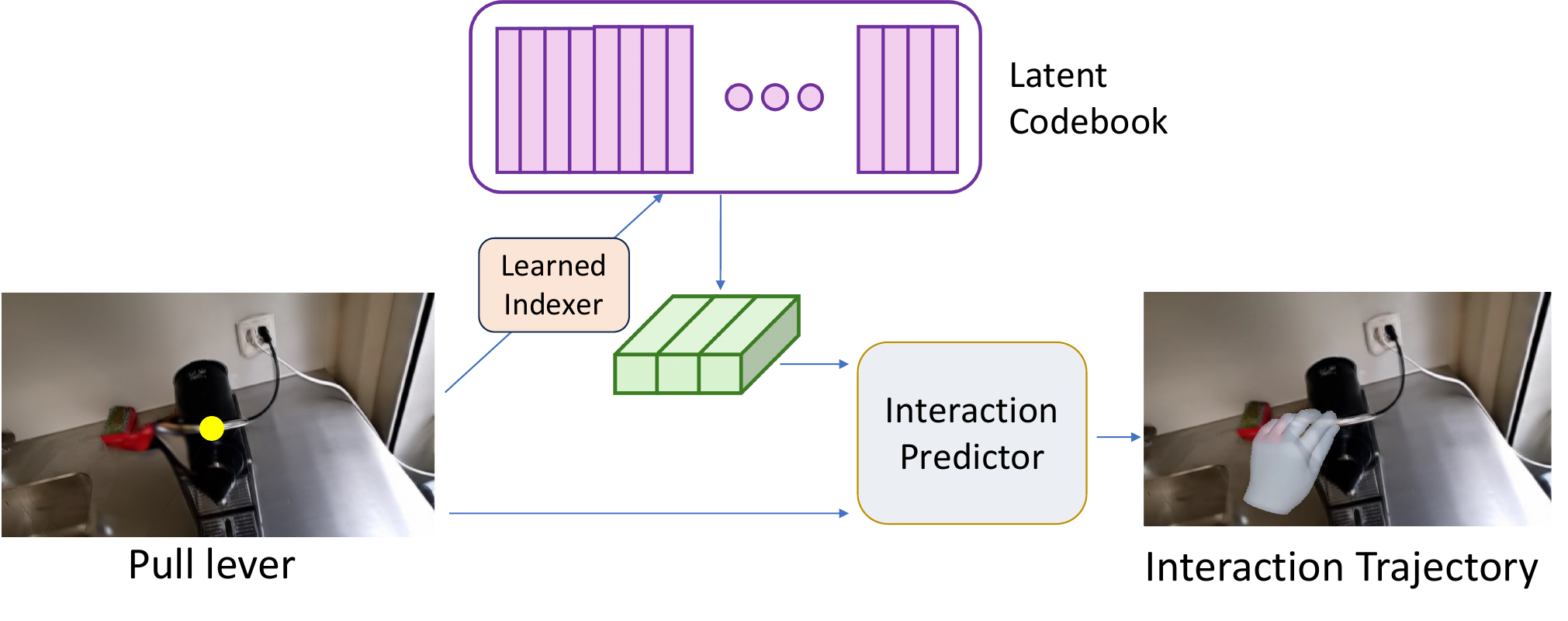}
        \caption{We train an Indexer module to map the inputs at test time to the codebook indices to extract the relevant embeddings. These are then passed to a interaction predictor module that outputs the 3D interaction trajectories. We consider 2 settings: (1) Forecasting: text describing the action, single image \& 3D contact point, (2) Interpolation: also providing the goal image showing the final state of the interaction (not shown here for clarity).
        }
        \figlabel{fig:predictor}
    \end{subfigure}
    \caption{{\bf Overview}. Our framework involves a 2-stage training procedure: (left) Interaction Codebook: to learn a latent codebook of hand poses and contact points, \ie, tokenizing interaction trajectories, (right) a learned Indexer \& an Interaction Predictor module to predict the interaction trajectories from single image, action text \& 3D contact point. We use pretrained features for images (from DeiT~\cite{Touvron2021ICML}) and text (from CLIP~\cite{Radford2021ICML}). 3D contact point is input as a 3D gaussian heatmap in a 3D voxel grid (omitted here for clarity).}
    \vspace{-0.3cm}
\end{figure*}

\boldparagraph{Hand mesh}
We use the MANO~\cite{Romero2017TOG} model to represent the hand. It consists of pose $\theta$ and shape $\beta$ parameters, which can be converted to a 3D hand mesh. These mesh vertices lie in the MANO coordinate system, which can be transformed to the \textit{camera coordinate system at the first timestep (reference frame)} using a rotation and translation. This gives us the hand trajectory in the reference frame.

\boldparagraph{Contact map}
We represent the contact points as a binary mask over the MANO hand mesh vertices, \ie, a 778 dimensional binary vector for each timestep, referred to as contact map in the paper (\Figref{fig:coordinate_frame}). The contact points may also change over time depending on the action, \eg twisting.

Once the 3D hand mesh is transformed into reference frame, we estimate the trajectory of the contact points by computing the centroid of the contact points at each timestep.
The contact centroid in the reference frame is used as the 3D contact point input during training. We assume a 3D contact point to be available in the first frame at test time. To input this point, we create a 3D voxel grid of resolution $16\times16\times16$. We take minimum and maximum values of 3D hand trajectory (along each axis, in the reference frame) across all training sequences. This span is divided into 16 equal parts per axis, associating each voxel to 3D metric range. It is kept fixed at train \& test time. Thus, the center of each voxel represents a 3D location in metric space, \ie a 3D metric grid, in the camera coordinate system. We create a 3D gaussian heatmap centered at 3D contact point in this grid, referred to as voxelized contact points in the paper. This changes as the trajectory evolves. This is different than the contact map (binary mask over mesh vertices).

\subsection{Interaction Codebook (InterCode)}
\seclabel{sec:latent}
This is implemented as a VQVAE with a transformer encoder-decoder architecture~\cite{Guo2024CVPR}. It consists of 3 modules:

\boldparagraph{Encoder}
The hand and contact map trajectories for a fixed time horizon T are passed to a transformer encoder to get the feature embedding. These features are used to query the codebook to obtain the relevant embeddings using L2 distance.  Specifically, the input trajectory, consisting of MANO parameters \& contact maps, is first passed to a 1-layer MLP to match the feature dimension of the transformer. We use a 1-layer transformer encoder with 1 head \& feature dimension $=512$. It outputs a $T\times512$ dimensional feature.

\boldparagraph{Latent codebook}
It is initialized as a $K \times E$ dimensional matrix and is updated during training using exponential moving average of the embeddings output from the encoder, following recent works~\cite{Guo2024CVPR}.  Concretely, we use a multi-layer residual VQVAE architecture~\cite{Guo2024CVPR}. We set the number of quantizers for VQVAE to 6, $K = 512$ and $E = 512$. When querying the codebook during training, sampling is done using the Gumbel-SoftMax trick~\cite{Jang2017ICLR} to preserve differentiability. We set the temperature of Gumbel-Softmax to 0.5 \& inject noise at training for stochastic sampling. At test time, argmax is used to get the closest codebook entry.

\boldparagraph{Decoder}
It takes multiple inputs: (1) the latent embeddings queried from the codebook ($T \times 512$), (2) CLIP~\cite{Radford2021ICML} embeddings of the text describing the action (512 dimensional feature extended for T timesteps), (3) DeiT~\cite{Touvron2021ICML} embeddings of the video showing the actions ($T \times 768$), (4) voxelized contact points at each timestep, which are passed through four 3D convolution layers. We also pass in a 4D grid with each voxel containing the 3D location of its center in metric space, and process with 3D convolution layers. This gives us a 32 dimensional feature vector from the contact module. Concatenating all these input features, we get a 1824-dimensional joint feature that is passed through a 1-layer MLP to change the dimensionality to 512 and then fed to a transformer decoder (1-layer with 1 head \& dropout of 0.2). The query embeddings for the transformer decoder are set as trainable parameters. It outputs a $T \times 512$ feature map which is passed to separate decoders for predicting the MANO parameters and the contact map. We use a 2-layer MLP to predict the MANO parameters and a 3-layer MLP to predict the contact maps for the entire time horizon $T$.

\boldparagraph{Training}
The loss function consists of several terms: (1) Smooth L1 loss on the MANO parameters, (2) L1 loss on the contact centroid, (3) L1 loss on the translation between the future frame and first frame, (4) L1 loss on the 6D rotation representation between the future frame and first frame, (5) binary cross-entropy loss on the contact map predictions.

\subsection{Learned Indexer}
\seclabel{sec:indexer}
At test time, we need to query the learned codebook to get the latent interaction embeddings. This is done using a learned module which takes the test time inputs, i.e. text describing an action, image showing an object \& a 3D contact point, and outputs a probability distribution over the codebook indices. The indices and corresponding embeddings can be sampled using the predicted probabilities. Note that the test time inputs only consist of a single image and a 3D contact point, unlike the entire trajectory used for training InterCode.

Specifically, we first compute features from different inputs: CLIP embedding of text (512), DeiT embeddings of images (768), features from voxelized contact point, in the reference frame, after passing through 4 3D convolutional layers (32) and the 3D contact point. We concatenate these features and process through a 2-layer MLP (1024 \& 512 nodes) to match the input dimensions of the transformer classifier module. It is implemented as a 1 layer, 1 head transformer decoder layer that takes in learnable query embeddings and aforementioned features from different modalities to output $T \times 512$ dimensional features. These features are then passed to a 3-layer MLP to output a probability distribution over the indices of the codebook.

\boldparagraph{Training}
It consists of a cross-entropy loss with ground truth computed as the closest codebook indices (L2 distance) from the encoder features of the InterCode module.

\subsection{Interaction Predictor (InterPred)}
\seclabel{sec:predictor}
Since the inputs at test time differ from the decoder inputs of the InterCode module during training, we need a separate learned module to predict the interaction trajectories. It consists of a transformer decoder module with the following inputs: (1) the latent embeddings queried from the codebook using the learned prior, (2) CLIP embeddings of the text describing the action, (3) DeiT embeddings of the image showing the object, and (4) voxelized contact point in the reference frame. These inputs are processed in the same way as for Learned Indexer before passing to the transformer decoder. The architecture is same as the decoder in InterCode. While the inputs are different than InterCode, they are passed to an MLP to match transformer dimension.

\boldparagraph{Training}
The loss function consists of several terms: (1) Smooth L1 loss on the MANO parameters, (2) L1 loss on the contact centroid, (3) L1 loss on the translation between the future frame and first frame, (4) L1 loss on the 6D rotation representation between the future frame and first frame, (5) binary cross-entropy loss on the contact map predictions.

\begin{figure}
    \centering
    \includegraphics[width=\columnwidth]{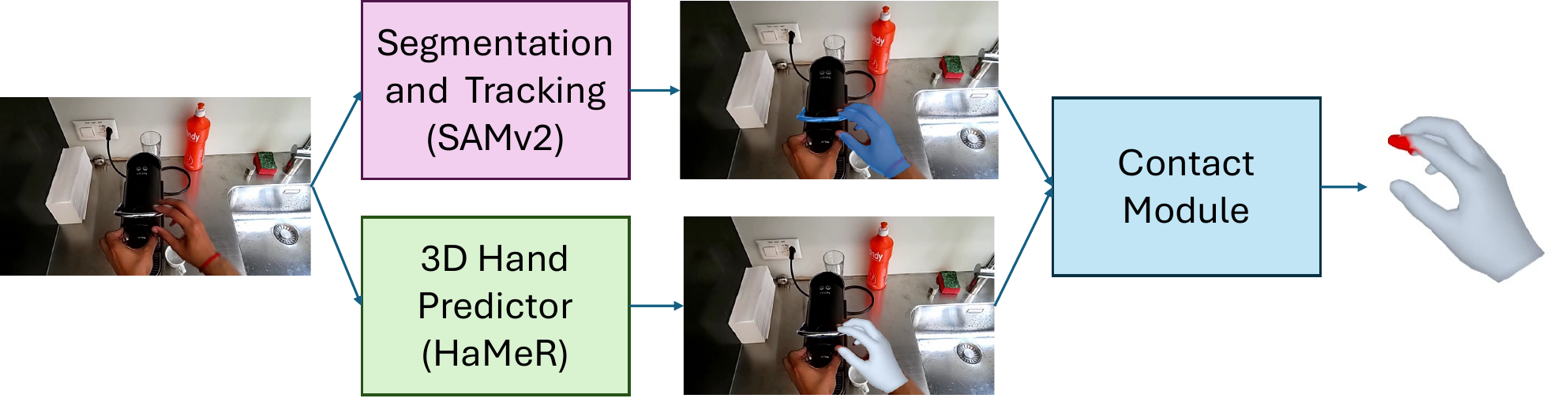}
    \includegraphics[width=\columnwidth]{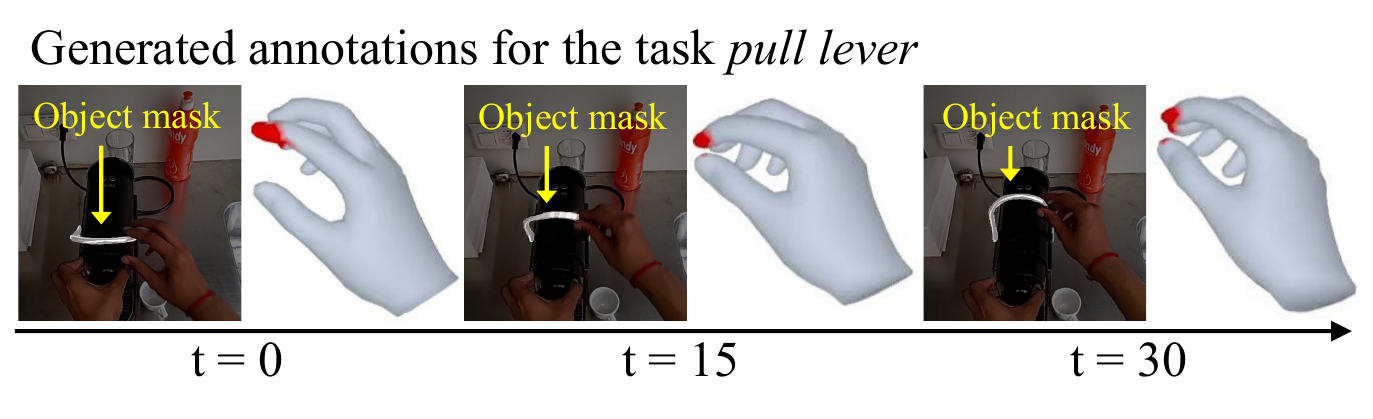}
    \vspace{-0.7cm}
    \caption{\textbf{Data engine}. (\textit{top})
    Object masks are extracted using SAMv2~\cite{Ravi2024ARXIV}, 3D hand poses \& masks (2D rendering of mesh) are from HaMeR~\cite{Pavlakos2024CVPR}, contact points are computed by projecting the 3D hand points into the 2D contact region (intersection of hand \& object masks). (\textit{bottom}) Generated object masks (highlighted in white), 3D hand mesh \& {\color{red}{contact points}} for 3 timesteps.}
    \figlabel{data_pipeline}
    \vspace{-0.3cm}
\end{figure}

\subsection{Data Engine}
\seclabel{sec:data}
Existing lab datasets are limited to simple interactions in constrained setups. Instead, we use egocentric videos from \holo showing everyday interactions with rigid, articulated \& deformable objects, \eg operating machines, tool use, kitchen tasks. It consists of 2k+ videos with temporally localized atomic actions, e.g. 
\textit{grab, open, screw, mix, rotate, align, slide, \etc} with diverse objects. We extract these clips of atomic actions and convert them into the required format for training \& evaluation. We consider 24 action categories with 120 object categories spanning 800 tasks. This is 2.5-10$\times$ larger than existing datasets, \eg HOT3D~\cite{Banerjee2024ARXIV} (33 rigid objects), ARCTIC~\cite{Fan2023CVPR} (10 articulated objects), GRAB~\cite{taheri2020grab} (51 objects), HOI4D~\cite{Liu2022CVPR} (16 object categories), in terms of objects \& diversity of interactions. While \holo contains diverse interactions, it does not provide 3D annotations for hand poses \& contact points. To extract these annotations, we design a semi-automatic data pipeline (\Figref{data_pipeline}) using 2D segmentation masks, 3D hand poses \& 2D contact region.

\boldparagraph{Segmentation masks}
We adopt a semi-automatic approach for object masks, that involves clicking a point on the target object in the first frame of a video \& tracking the object mask across the video using SAMv2~\cite{Ravi2024ARXIV}. For 2D hand masks, we render the 3D hand mesh (described next) into the image.

\begin{table*}
    \centering
    \setlength{\tabcolsep}{2.5pt}
    \resizebox{\linewidth}{!}{
    \begin{tabular}{c l c c c c c c c c c c c c}
        \toprule
        & Method & \multicolumn{3}{c}{Task-level} & \multicolumn{3}{c}{Object-level} & \multicolumn{3}{c}{Action-level} & \multicolumn{3}{c}{Scene-level} \\
        \cmidrule(lr){3-5}
        \cmidrule(lr){6-8}
        \cmidrule(lr){9-11}
        \cmidrule(lr){12-14}
        & & M-PE(cm)$\downarrow$ & M-PA(cm)$\downarrow$ & F1$\uparrow$ & M-PE(cm)$\downarrow$ & M-PA(cm)$\downarrow$ & F1$\uparrow$ & M-PE(cm)$\downarrow$ & M-PA(cm)$\downarrow$ & F1$\uparrow$ & M-PE(cm)$\downarrow$ & M-PA(cm)$\downarrow$ & F1$\uparrow$ \\
        \midrule
        {\multirow{8}{*}{\rotatebox{90}{Hand visible}}} & \bf Forecasting & & & & & & & & & & & & \\
        & \quad \ctf & 8.32 & 3.06 & 0.72 & 8.70 & 3.19 & 0.72 & 8.59 & 3.25 & 0.73 & 8.52 & 3.09 & 0.72 \\
        & \quad \cdiff & 8.42 & \bf 2.75 & 0.72 & 9.01 & \bf 2.88 & 0.73 & 9.11 & \bf 3.10 & 0.74 & 9.49 & \bf 2.84 & 0.71 \\
        & \quad \ltf (Ours) & \bf 7.61 & 2.99 & \bf 0.75 & \bf 8.09 & 3.06 & \bf 0.75 & \bf 8.14 & 3.32 & \bf 0.77 & \bf 7.87 & 3.04 & \bf 0.74 \\
        & \bf Interpolation & & & & & & & & & & & & \\
        & \quad \ctf & 7.52 & 3.02 & 0.76 & 8.10 & 3.10 & 0.76 & 7.77 & 3.29 & 0.78 & 7.54 & 3.00 & 0.76 \\
        & \quad \cdiff & 8.30 & \bf 2.72 & 0.77 & 8.52 & \bf 2.91 & \bf 0.78 & 8.38 & \bf 3.16 & 0.78 & 8.60 & \bf 2.81 & 0.77 \\
        & \quad \ltf (Ours) & \bf 6.72 & 2.87 & \bf 0.80 & \bf 7.48 & 3.02 & \bf 0.78 & \bf 7.25 & 3.19 & \bf 0.80 & \bf 7.27 & 2.93 & \bf 0.78 \\
        \midrule
        {\multirow{8}{*}{\rotatebox{90}{No hands}}} & \bf Forecasting & & & & & & & & & & & & \\
        & \quad \ctf & 8.34 & 2.95 & 0.72 & 8.85 & 3.02 & 0.73 & 8.54 & 3.21 & 0.74 & 8.53 & 3.05 & 0.72 \\
        & \quad \cdiff & 8.85 & \bf 2.65 & 0.72 & 9.74 & \bf 2.81 & 0.72 & 9.50 & \bf 3.11 & 0.74 & 9.21 & \bf 2.74 & 0.71 \\
        & \quad \ltf (Ours) & \bf 7.93 & 2.97 & \bf 0.76 & \bf 8.26 & 3.01 & \bf 0.76 & \bf 8.19 & 3.28 & \bf 0.77 & \bf 7.90 & 3.00 & \bf 0.75 \\
        & \bf Interpolation & & & & & & & & & & & & \\
        & \quad \ctf & 7.30 & 2.95 & 0.76 & 8.42 & 3.15 & 0.76 & 8.01 & 3.30 & 0.77 & 7.29 & 2.99 & 0.77 \\
        & \quad \cdiff & 8.27 & \bf 2.76 & 0.76 & 8.68 & \bf 2.91 & 0.76 & 9.33 & \bf 3.10 & 0.78 & 10.02 & \bf 2.82 & 0.76 \\
        & \quad \ltf (Ours) & \bf 7.13 & 2.89 & \bf 0.79 & \bf 7.24 & 2.97 & \bf 0.79 & \bf 7.44 & 3.21 & \bf 0.80 & \bf 7.12 & 2.89 & \bf 0.78 \\
        \bottomrule
    \end{tabular}
    }
    \caption{{\bf Generalization results}. We report MPJPE (M-PE: cm), MPJPE-PA (M-PA: cm) \& F1-score in 4 generalization settings: novel tasks, objects, actions \& scene, to measure the accuracy of the predicted trajectories. We adapt two recent methods from human pose literature to work with image inputs, \ie, \ctf : ViT encoded image features passed to a transformer decoder (similar to pose decoder of T2P~\cite{Jeong2024CVPR}), \cdiff : MDM~\cite{Tevet2023ICLR} modified to take image features as well. \ltf leads to better absolute hand poses \& contact maps.
    }
    \tablelabel{tab:results}
    \vspace{-0.3cm}
\end{table*}

\boldparagraph{3D hand poses}
We use the off-the-shelf HaMeR~\cite{Pavlakos2024CVPR} model to get 3D hand meshes in each video frame. HaMeR takes an image crop around the hand as input (estimted from Hands23 model~\cite{cheng2023towards}). The predicted 3D mesh in each frame is transformed to the reference frame using known intrinsics \& extrinsics available in the \holo dataset.

\boldparagraph{3D Contact points}
We first compute a 2D contact region in an image as the overlap between the hand \& object masks. We add gaussian noise at the boundary to obtain a dense 2D contact region. We back project the 2D contact region into 3D hand mesh to obtain 3D contact points on the hand mesh.

\section{Experiments}
\seclabel{sec:experiments}

\subsection{Protocol}

\boldparagraph{Task}
We predict trajectories of MANO hand parameters \& contact map (a binary mask over $778$ vertices of the MANO~\cite{romero2017embodied} hand mesh) in 2 settings: (1) Forecasting: input consists of a textual description of the action, a current image showing the object of interaction, and a 3D contact point. (2) Interpolation: We provide the goal image, along with the previous inputs, showing the final state of interaction. For each setting, we 2 two variants: (a) current image with a hand, (b) current image without a hand (removed via in-painting~\cite{Ye2023ICCV}) since the hand may not be visible before the start of the interaction in many practical scenarios. We only consider right hand motion in this work.

\boldparagraph{Generalization aspects}
We study generalization capabilities for $4$ different settings: (1) Object categories, (2) Action categories, (3) Novel tasks: a combination of object \& action categories, (4) Novel scene: we hold out videos from 1 location (data is collected in 2 locations).
For each setting, we create train, validation \& test splits in the ratio 80:10:10.

\boldparagraph{Metrics}
(1) MPJPE: Mean Per Joint Position Error between predicted \& ground truth hand joints, averaged over all future timesteps. (2) MPJPE-PA is the procrustus aligned variant, which optimizes for a single rotation, translation \& scale to align the entire trajectory with ground truth trajectory, before computing MPJPE.
(3) F1 score: harmonic mean of precision \& recall to measure accuracy of contact maps .

\boldparagraph{Baselines}
Since no prior works predict interaction trajectories from image input, we adapt 2 recent baselines from human pose literature to work in our setting: (1) \ctf: The input image is encoded using a ViT~\cite{Dosovitskiy2021ICLR} encoder followed by a transformer decoder, similar to the pose decoder used in T2P~\cite{Jeong2024CVPR}. TCP decoder takes in trajectory query (for global intent), pose query \& pose embeddings. In our case, intent comes from text. We replace pose embeddings with \{image, contact, text\} features, pass them to an MLP \& then feed to \{TRM, MLP\} modules, as in TCP. (2) \cdiff: We modify the motion diffusion model (MDM~\cite{Tevet2023ICLR}) for hand trajectory and contact map prediction. For this, we change conditioning to use \{image, text, contact\} features but diffusion architecture is same. Both \ctf and \cdiff are trained in the same setting as ours for fair comparisons. 

Existing methods that use object-centric representations~\cite{Christen2024ARXIV,Cha2024CVPR} can not be applied in our setting since we do not have the 3D object models. Hand-centric representations~\cite{Ye2024CVPR,Taheri2024THREEDV} are also not suitable in our setting since the hand is not always visible in the input image at test time.

Our approach involves a 2-stage training procedure: InterCode followed by Indexer \& InterPred, from different inputs. We train with both transformer \& diffusion variants for InterPred, referred to as \ltf \& \ldiff respectively. \ldiff uses same InterCode \& Indexer as \ltf but different InterPred, \ie, an iterative denoising model from MDM instead of a single step prediction. The denoising is done in latent space of the codebook followed by a decoder to generate trajectories. It is trained using L2 loss on denoised embedding and loss in~\Secref{sec:predictor}.

\subsection{Results}
\seclabel{sec:results}

\boldparagraph{Generalization results}
First, we study if the model can generate trajectories from the text describing the action, single image \& 3D contact point, \ie, Forecasting task. In the absence of a goal, this is quite challenging and inherently multimodal in nature. We compare with the GT trajectories for all 4 generalization aspects: novel objects, actions, tasks \& scene. We report MPJPE, MPJPE-PA for hand trajectories \& F1-score for contact maps. We observe that our approach leads to consistent gains in absolute hand poses \& contact maps. We show results (\Tableref{tab:results}) on 2 variants of the task, (1) hand visible in the image, (2) hand absent in the image, often the case at the start of interactions. The hand visible in the image makes the tasks slightly easier since the initial pose is visible, as also evident from the results.

Next, we also provide the goal image, along with the previous inputs, showing the final state of interaction, referred to as goal-conditioned interpolation. This is slightly easier than the Forecasting task since the final state of the interaction is also provided. The results (\Tableref{tab:results}) on 2 varaints of the task, (1) hand visible in the image, (2) hand absent in the image, show similar benefits as in Forecasting. 

These results are for $T$=30 timesteps. We also report results for $T$=16 in the supplementary with similar trends.

\begin{table}
    \centering
    \setlength{\tabcolsep}{1pt}
    \resizebox{1.0\linewidth}{!}
    {
    \begin{tabular}{>{\raggedright\arraybackslash}p{6.2cm} c c c}
        \toprule
        Method & MPJPE$\downarrow$ & MPJPE-PA$\downarrow$ & F1$\uparrow$ \\
        \midrule
        \ctf: trained on ARCTIC & 15.76 & 3.61 & 0.36 \\
        \ltf: trained on ARCTIC & \bf 14.77 & 3.76 & 0.41 \\
        \ltf: zero-shot from Holo & 15.72 & 3.71 & 0.19 \\
        \ltf: Codebook \& Indexer from Holo & 15.36 & \bf 3.58 & \bf 0.45 \\
        \bottomrule
    \end{tabular}
    }
    \caption{HoloAssist is significantly larger than ARCTIC in terms of contact sequences leading to benefits in zero-shot generalization \& transferring models trained on HoloAssist to ARCTIC.}
    \tablelabel{tab:cross_dataset}
    \vspace{-0.5cm}
\end{table}

\boldparagraph{Results on ARCTIC}
We also show results (\Tableref{tab:cross_dataset}) on ARCTIC~\cite{Fan2023CVPR} to check if \ltf leads to gains on existing lab datasets. ARCTIC has accurate 3D labels but is limited in scale (10$\times$ smaller than HoloAssist for contact sequences). When trained only on ARCTIC, we find the trends to transfer from HoloAssist, \ie, \ltf is better than \ctf. Using codebook \& indexer trained on HoloAssist further benefits models trained on ARCTIC. Zero-shot transfer from HoloAssist is competitive to models trained on ARCTIC.

\begin{table*}
    \centering
    \setlength{\tabcolsep}{2.5pt}
    \resizebox{\linewidth}{!}{
    \begin{tabular}{c l c c c c c c c c c c c c}
        \toprule
        & Method & \multicolumn{3}{c}{Task-level} & \multicolumn{3}{c}{Object-level} & \multicolumn{3}{c}{Action-level} & \multicolumn{3}{c}{Scene-level} \\
        \cmidrule(lr){3-5}
        \cmidrule(lr){6-8}
        \cmidrule(lr){9-11}
        \cmidrule(lr){12-14}
        & & M-PE(cm)$\downarrow$ & M-PA(cm)$\downarrow$ & F1$\uparrow$ & M-PE(cm)$\downarrow$ & M-PA(cm)$\downarrow$ & F1$\uparrow$ & M-PE(cm)$\downarrow$ & M-PA(cm)$\downarrow$ & F1$\uparrow$ & M-PE(cm)$\downarrow$ & M-PA(cm)$\downarrow$ & F1$\uparrow$ \\
        \midrule
        {\multirow{6}{*}{\rotatebox{90}{Hand visible}}} & \bf Forecasting & & & & & & & & & & & & \\
        & \quad \ltf & 7.61 & 2.99 & 0.75 & 8.09 & 3.06 & 0.75 & 8.14 & 3.32 & 0.77 & 7.87 & 3.04 & 0.74 \\
        & \quad \ldiff & 7.90 & 3.03 & 0.71 & 8.49 & 3.20 & 0.72 & 8.29 & 3.34 & 0.72 & 8.32 & 3.10 & 0.72\\
        & \bf Interpolation & & & & & & & & & & & & \\
        & \quad \ltf & 6.72 & 2.87 & 0.80 & 7.48 & 3.02 & 0.78 & 7.25 & 3.19 & 0.80 & 7.27 & 2.93 & 0.78 \\
        & \quad \ldiff & 6.53 & 2.62 & 0.78 & 7.23 & 2.81 & 0.78 & 7.11 & 3.10 & 0.79 & 6.70 & 2.84 & 0.78 \\
        \midrule
        {\multirow{6}{*}{\rotatebox{90}{No hands}}} & \bf Forecasting & & & & & & & & & & & & \\
        & \quad \ltf & 7.93 & 2.97 & 0.76 & 8.26 & 3.01 & 0.76 & 8.19 & 3.28 & 0.77 & 7.90 & 3.00 & 0.75 \\
        & \quad \ldiff & 7.82 & 2.89 & 0.71 & 8.63 & 3.06 & 0.71 & 8.19 & 3.15 & 0.72 & 7.93 & 2.96 & 0.73 \\
        & \bf Interpolation & & & & & & & & & & & & \\
        & \quad \ltf & 7.13 & 2.89 & 0.79 & 7.24 & 2.97 & 0.79 & 7.44 & 3.21 & 0.80 & 7.12 & 2.89 & 0.78 \\
        & \quad \ldiff & 6.60 & 2.65 & 0.76 & 7.73 & 2.78 & 0.76 & 7.42 & 3.04 & 0.78 & 6.89 & 2.71 & 0.77 \\
        \bottomrule
    \end{tabular}
    }
    \caption{
    {\bf \ldiff trends}. Training InterPred with diffusion loss leads to better hand poses but worse contact maps than \ltf.
    }
    \tablelabel{tab:results_diff}
    \vspace{-0.3cm}
\end{table*}

\boldparagraph{Benefits of Interaction Codebook}
We verify if the 2-stage training procedure is beneficial by comparing with single-stage training methods \ctf \& \cdiff in~\Tableref{tab:LatentAct} and observe consistent benefits. Another option is to retrieve an interaction trajectory from the training set, conditioned on the input. However, this does not scale well with the training data \& is computationally expensive, \eg the learned codebook in InterCode has 512 entries whereas the training dataset has $\sim$15K sequences, which is significantly larger.

\boldparagraph{Contact maps helps hand predictions}
Here, we train the InterCode \& InterPred modules with only the hand pose loss, i.e., removing the loss on contact maps. In~\Tableref{tab:contact_map}, we observe that loss on contact maps helps with hand predictions.

\begin{table}
    \centering
    \resizebox{0.95\linewidth}{!}
    {
    \begin{tabular}{l c c c}
        \toprule
        Method & MPJPE$\downarrow$ & MPJPE-PA$\downarrow$ & F1$\uparrow$ \\
        \midrule
        \ctf & 7.52 & 3.02 & 0.76\\
        \ctf + InterCode & 6.72 & 2.87 & 0.80\\
        \cdiff & 8.30 & 2.72 & 0.77\\
        \cdiff + InterCode & 6.53 & 2.62 & 0.78\\
        \bottomrule
    \end{tabular}
    }
    \caption{Two stage training with InterCode improves over single-stage training methods \ctf \& \cdiff. Note that \ctf + InterCode is same as \ltf. 
    }
    \tablelabel{tab:LatentAct}
    \vspace{-0.2cm}
\end{table}

\boldparagraph{Trends with \ldiff}. Training InterPred with diffusion losses leads to better hand poses but worse contact maps compared to \ltf, across all settings (\Tableref{tab:results_diff}).

\boldparagraph{Dataset scale} In~\Figref{fig:scale}, we see that both \ltf \& \ctf improve with larger training dataset size.

\begin{figure}
    \centering
    \includegraphics[width=\columnwidth]{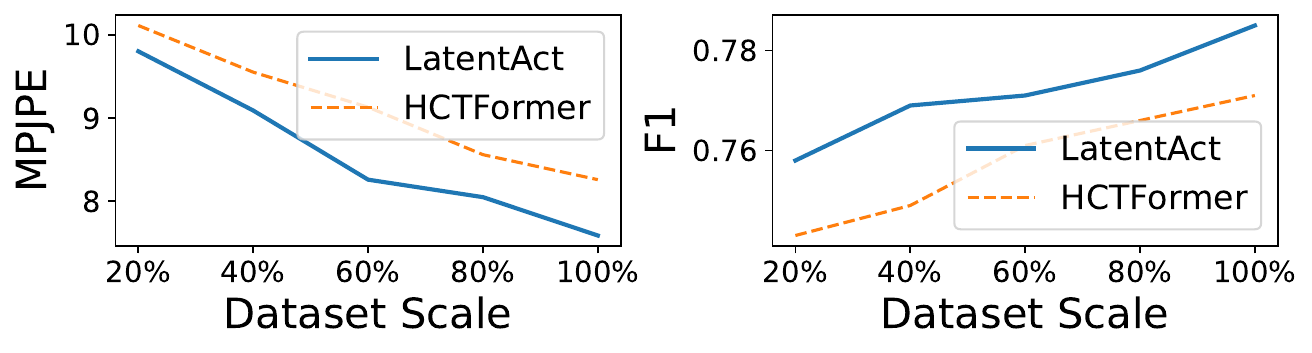}
    \caption{More training data helps both \ltf \& \ctf.
    }
    \figlabel{fig:scale}
    \vspace{-0.5cm}
\end{figure}

\subsection{Visualizations}

\begin{figure*}
    \centering
    \resizebox{0.95\linewidth}{!}
    {
    \renewcommand{\arraystretch}{0} %
    \setlength{\lightrulewidth}{0.1pt}  %
    \setlength{\heavyrulewidth}{0.1pt}
    \begin{tabular}{l}
        \toprule
        \includegraphics[width=\linewidth]{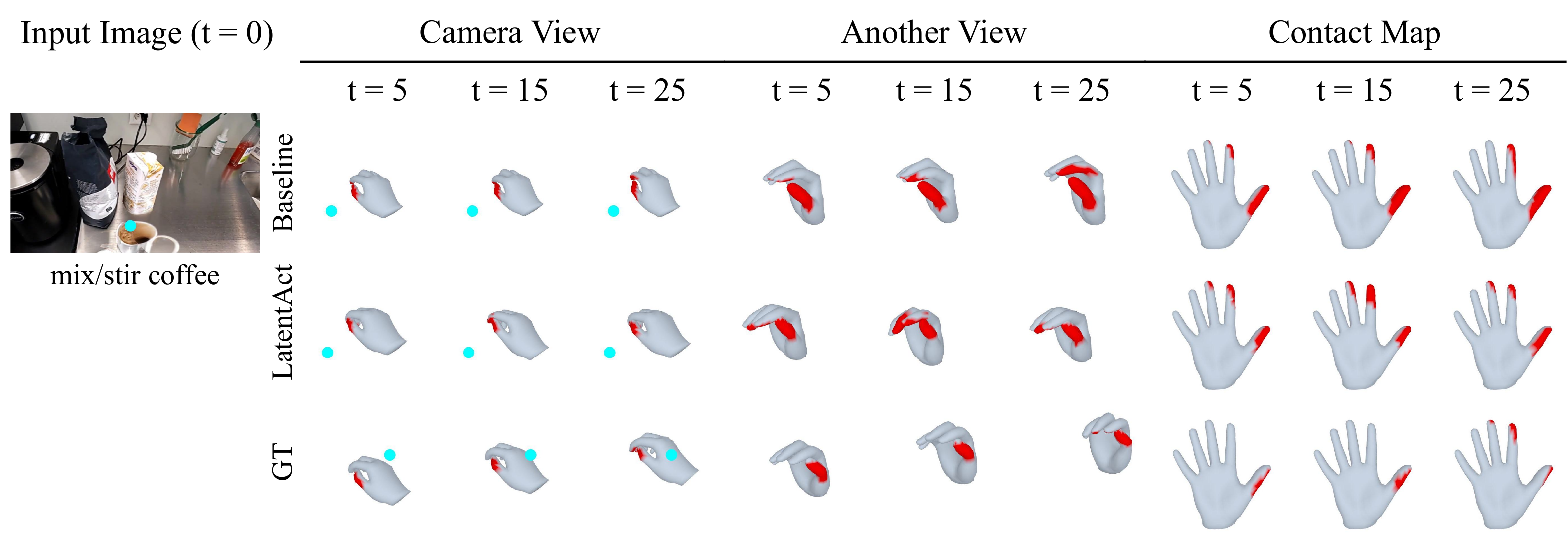} \\
        \midrule
        \includegraphics[width=\linewidth]{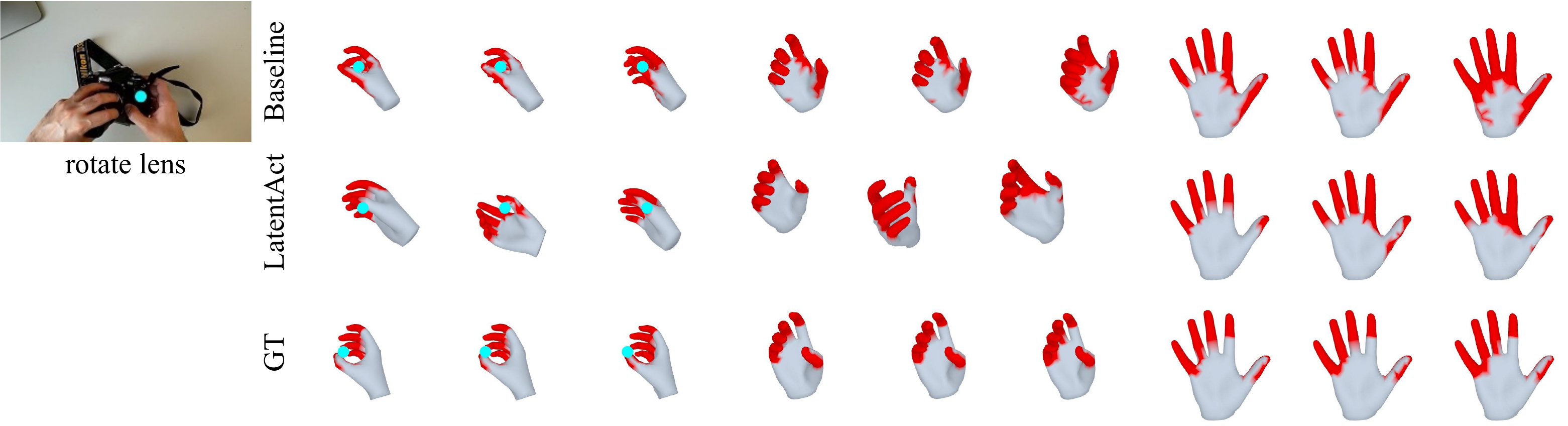} \\
        \midrule
        \includegraphics[width=\linewidth]{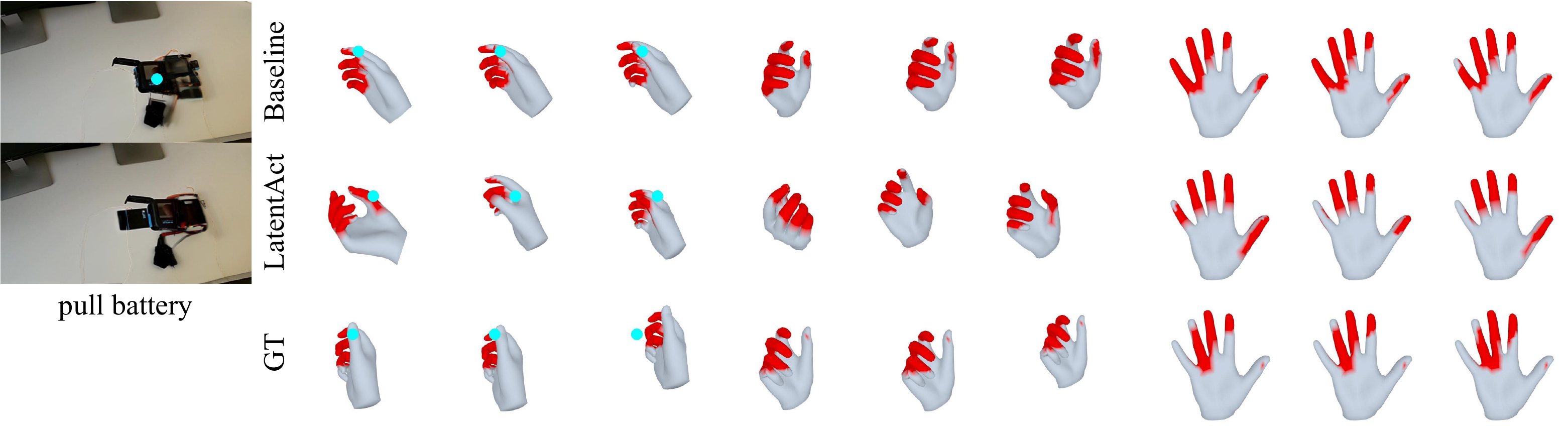} \\
        \midrule
        \includegraphics[width=\linewidth]{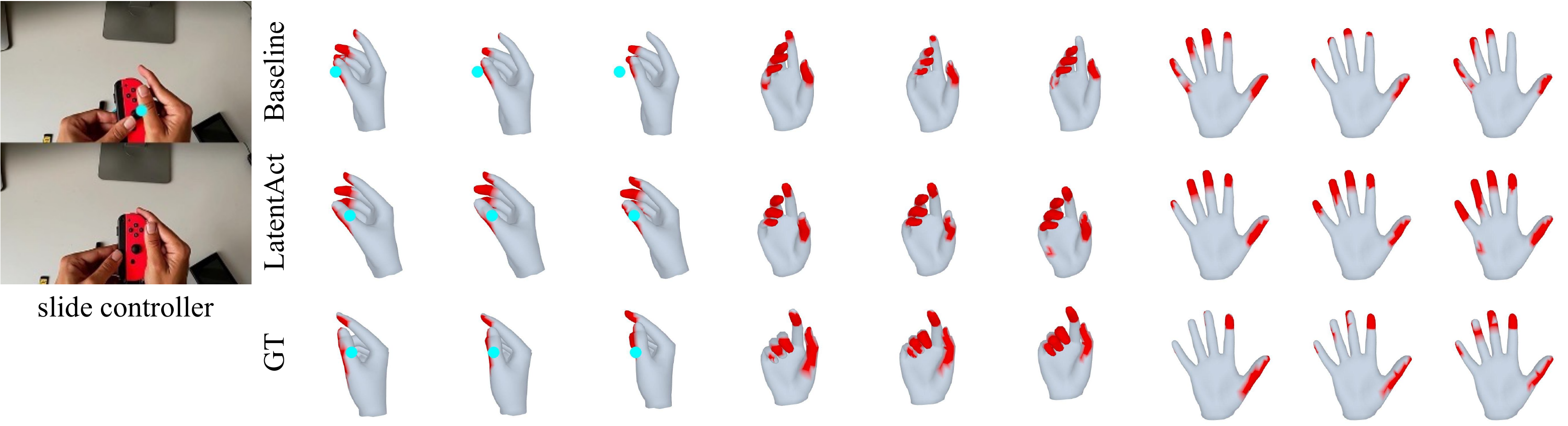} \\
        \bottomrule
    \end{tabular}   
    }
    \caption{\textbf{Visualizations}. We compare the predictions of \ltf \& the best baseline on a few examples from the task level generalization setting for both forecasting \& interpolation for 3 timesteps (t=5, 15, 25). The left column shows the input image with the contact point (projected in the image in {\color{cyan}{cyan blob}}), goal image (for interpolation task) \& text describing the action. The other columns show: (a) Camera View: predicted hand in the camera view with the contact point, this captures the placement of the hand around the contact point, (b) Another View: this better visualizes the hand pose from a different camera viewpoint, (c) Contact Maps are shown as {\color{red}{red parts}} of the hand mesh. \ltf leads to better orientation of hand \& sharper contact maps than the baseline. \textit{Note that these tasks are not seen during training.
    }}
    \figlabel{fig:viz}
    \vspace{-0.5cm}
\end{figure*}

We compare the predictions of \ltf \& the best baseline on a few examples from task level generalization setting (\Figref{fig:viz}) for both forecasting \& interpolation for 3 timesteps (t=5, 15, 25). The left column shows the input image with the contact point (projected in the image in {\color{cyan}{cyan blob}}), goal image (for interpolation task) \& text describing the action. The other columns show: (a) Camera View: predicted hand in the camera view with the contact point, this captures the placement of the hand around the contact point, (b) Another View: this better visualizes the hand articulation from a different camera viewpoint, (c) Contact Maps are shown as {\color{red}{red parts}} of the hand mesh. \ltf leads to better orientation of hand \& sharper contact maps than the baseline. Note that these tasks are not seen during training.
More visualizations \& analysis are provided in the supplementary.

\begin{table}
    \centering
    \resizebox{0.95\linewidth}{!}{
    \begin{tabular}{l c c c}
        \toprule
        Method & MPJPE$\downarrow$ & MPJPE-PA$\downarrow$ & F1$\uparrow$ \\
        \midrule
        \ctf & 7.52 & 3.02 & 0.76 \\
        \quad $-$ no contact map & 7.65 & 2.89 & -- \\
        \ltf & 6.72 & 2.87 & 0.80 \\
        \quad $-$ no contact map & 7.76 & 2.91 & -- \\
        \cdiff & 8.30 & 2.72 & 0.77 \\
        \quad $-$ no contact map & 8.43 & 2.78 & -- \\
        \ldiff & 6.53 & 2.62 & 0.78 \\
        \quad $-$ no contact map & 7.09 & 2.90 & -- \\
        \bottomrule
    \end{tabular}
    }
    \caption{Loss on contact maps helps hand predictions for all models.}
    \tablelabel{tab:contact_map}
    \vspace{-0.3cm}
\end{table}

\boldparagraph{Limitations}
We do not predict the object state change. While 3D object models are not available, alternatives like predicting 2D object masks could be explored. While we assume a 3D contact point to be available, it can also be estimated using off-the-shelf depth models \& ground truth intrinsics to get the 3D location of a 2D point that can be prompted by a user (\eg by clicking a point on the image). Our experiments measure whether the predicted interaction trajectory is accurate, but it does not account for the multimodal nature of future prediction. The annotations in our work are programatically generated and may not be accurate in all cases. While we filter out the erroneous cases, using a better annotation tool~\cite{Tripathi2023ICCV} would be beneficial.

\section{Conclusion}
\label{sec:conclusion}

Our model predicts interaction trajectories, \ie 3D hand motion \& contact maps, from single RGB view, action text \& 3D contact point as input. It consists of: (1) Interaction Codebook: a VQVAE to learn a latent codebook of 3D hand poses \& contact points, \ie, tokenizing interactions. (2) Interaction Predictor: a transformer-decoder module to predict the interaction trajectory from test time inputs, by using an indexer module to retrieve a latent interaction from the learned codebook. Our large-scale experiments on the diverse HoloAssist dataset show benefits across 4 generalization settings. Our work is a first step towards learning a motion prior on interaction sequences and can be augmented with object trajectories with better 3D models in the future.
Our data pipeline is modular and can also incorporate better hand \& segmentation models in future.

\clearpage
\boldparagraph{Acknowledgments}
This material is based upon work supported by an NSF CAREER Award (IIS2143873), USDA-NIFA AIFARMS National AI Institute (USDA\#2020-67021-32799) and NSF (IIS-2007035). Part of the work was done when Aditya Prakash was an intern at Microsoft. We acknowledge compute support by NVIDIA Academic Hardware Grant and a DURIP grant. We thank Shaowei Liu and Arjun Gupta for feedback on the draft.

{
    \small
    \bibliographystyle{ieeenat_fullname}
    \bibliography{biblioLong, refs}
}

\end{document}